\pdfoutput=1
\documentclass[11pt]{article}
\usepackage{tabularx} 
\usepackage{booktabs}
\usepackage[svgnames,table]{xcolor}
\usepackage[tableposition=above]{caption}

\usepackage{acl}
\usepackage{macros}
\usepackage{rycolab}
\usepackage{linguex}
\usepackage{fontawesome}

\usepackage{times}
\usepackage{latexsym}

\usepackage[T1]{fontenc}
\usepackage[utf8]{inputenc}

\usepackage{microtype}

\usepackage{inconsolata}

\title{
Which Spurious Correlations Impact Reasoning in NLI Models?\\ A Visual Interactive Diagnosis through Data-Constrained Counterfactuals
}

\author{
Robin Chan$^{1}$~\;~Afra Amini$^{1, 2}$~\;~Mennatallah El-Assady$^{1, 2}$ \\
  $^1$ETH Z\"{u}rich~\;~$^2$ETH AI Center\\
 \texttt{\href{mailto:chanr@ethz.ch}{chanr@ethz.ch}}~\;~\texttt{\{\href{mailto:afra.amini@inf.ethz.ch}{afra.amini}, \href{mailto:melassady@inf.ethz.ch}{melassady}\}@inf.ethz.ch}
}

\begin{document}
\maketitle
\begin{abstract}
We present a human-in-the-loop dashboard tailored to diagnosing potential spurious features that NLI models rely on for predictions. 
The dashboard enables users to generate diverse and challenging examples by drawing inspiration from GPT-3 suggestions. 
Additionally, users can receive feedback from a trained NLI model on how challenging the newly created example is and make refinements based on the feedback.
Through our investigation, we discover several categories of spurious correlations that impact the reasoning of NLI models, which we group into three categories: Semantic Relevance, Logical Fallacies, and Bias. Based on our findings, we identify and describe various research opportunities, including diversifying training data and assessing NLI models' robustness by creating adversarial test suites. 

\vspace{0.7em}
\hspace{.5em}\raisebox{-0.055cm}{\includegraphics[width=1.7em,height=0.9em]{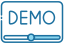}}\hspace{2em}\parbox{\dimexpr\linewidth-2\fboxsep-2\fboxrule}{\url{https://dcc.lingvis.io}}
%\vspace{1em}
%\hspace{1.25em}\faGithub \hspace{1em} \url{https://github.com/chanr0/dcc}
\end{abstract}
\begin{figure*}
    \centering
    \includegraphics[width=\textwidth]{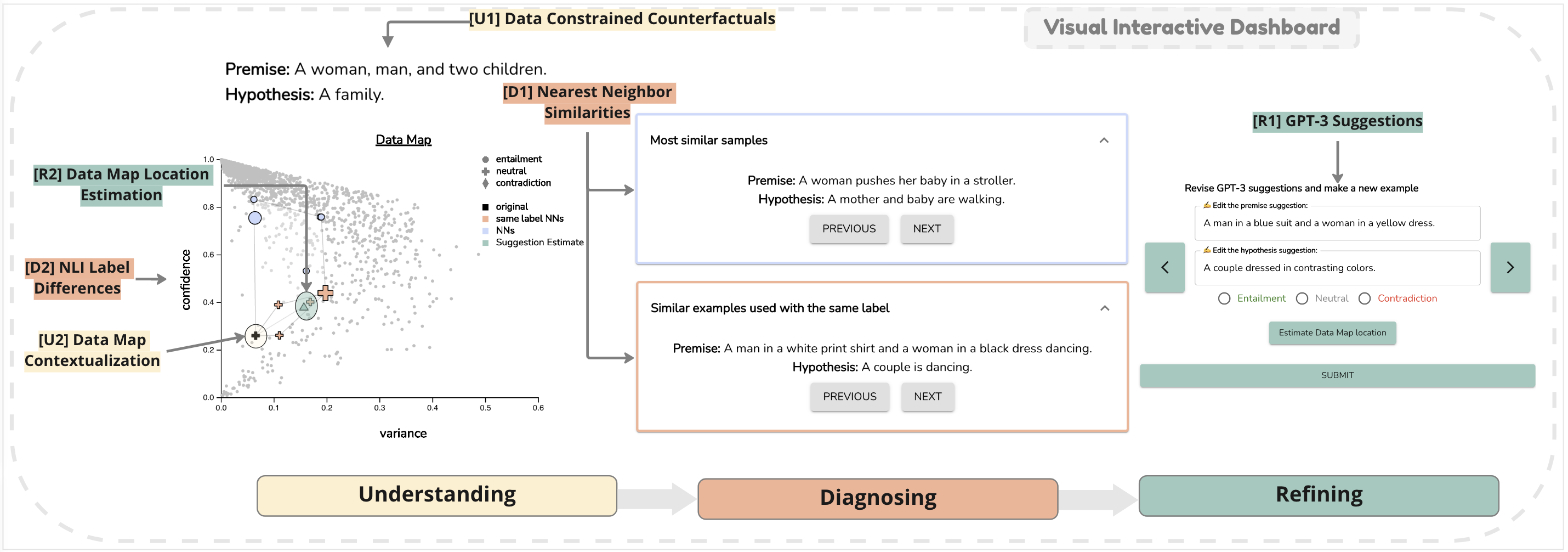}    \caption{The three main phases in our interactive dashboard. In the first step, \ui{1}, \ui{2}, the user understands the main data point and the prediction of the model on that data point. In the second step \di{1}, \di{2}, the user diagnoses the similarities and differences between other data points in the dataset and the main data point. In the last step, \ri{1}, \ri{2}, the user revises GPT-3 suggestions using the feedback from the model and submits a counterfactual.}
    \label{fig:dashboard}
\end{figure*}

\section{Introduction}
The availability of crowdsourced large-scale datasets has been influential in the field of natural language processing. These datasets have empowered advancements in a wide range of downstream tasks, including the natural language inference (NLI) task \citep[SNLI;][]{bowman-etal-2015-large}. While being influential, crowdsourcing frameworks can introduce artifacts, biases, and spurious correlations that can negatively impact the robustness and out-of-domain generalization of the models that are trained on such datasets \citep{jia-liang-2017-adversarial, mccoy-etal-2019-right}.

A \defn{spurious correlation} exists when a feature correlates with the target label while there is no causal relationship between the feature and the label. For example, the fact that a sentence includes the word ``amazing'' (as a feature) might correlate with a positive sentiment but does not \emph{cause} the sentiment label to be positive, as one can imagine crafting a sentence like ``the amazing concert was ruined by the terrible acoustics in the venue'', which has a negative sentiment. It has been shown that such spurious correlations exist in crowdsourced datasets \citep{gardner-etal-2021-competency}, and this will prevent models that are trained on these datasets from performing on adversarial or out-of-domain test sets \citep{mccoy-etal-2019-right}.

One approach to prevent a model from relying on spurious correlations between a feature and the label is to break such correlations by providing \defn{counterfactuals} during training. In this context, counterfactuals are data points that contain the feature but have a different label. Following our previous example, ``the amazing concert was ruined by the terrible acoustics in the venue'' is a counterfactual sentence since it contains the word ``amazing'' but has a negative sentiment. Augmenting datasets with counterfactuals can break the spurious correlations and help the model to generalize to out-of-domain examples. However, generating counterfactuals is challenging; it involves first identifying noncausal features that correlate with the label, i.e., spurious correlations, and then generating the counterfactuals for a given feature.  

One simple approach to generate counterfactuals is through minimal edits. In this approach, the first step---identifying spurious correlations---is bypassed. Therefore, counterfactuals are generated without targeting any specific feature. To generate a counterfactual, existing data points in the dataset are edited minimally such that they have a different label compared to their original one. While such an approach is scalable and can be effective in certain scenarios \citep{khashabi-etal-2020-bang}, creating counterfactuals through minimal edits does \emph{not} necessarily improve the generalization of models and might even hurt the performance \citep{huang-etal-2020-counterfactually}. Therefore, there is a need for a more nuanced and innovative approach to counterfactual generation.\looseness=-1

In this paper, we propose a data-centric approach to counterfactual generation. First, we identify existing counterfactuals in datasets, which we term \defn{data-constrained counterfactuals} (DCC). Second, using our interactive dashboard, we diagnose features that spuriously correlate with their label by comparing and contrasting the DCC with other data points in the dataset. Lastly, we generate a diverse set of counterfactual examples with the help of GPT-3 \citep[\texttt{davinci-003};][]{gpt3}. 

Overall, our dashboard offers a human-in-the-loop, or more generally, a mixed-initiative approach. A user can diagnose spurious correlations and common patterns that result in NLI models' inability to predict the labels correctly. Finding such weak spots can provide ways to improve the NLI model. Furthermore, after the user has generated a set of new counterfactuals, the NLI model can give feedback on how valuable each counterfactual is by expressing its uncertainty in predicting the sample's annotated label. This can help the user to revise the counterfactual and improve the usefulness of the generated set.\looseness=-1

While our dashboard can be extended to various tasks in natural language processing, we focus on the NLI task in this work. Using our dashboard, we find a variety of features that correlate spuriously with labels. We categorize these features into three categories, which we name: Semantic Relevance, Logical Fallacies, and Biases. We further find a category of samples that are annotation artifacts. Based on these findings, and with the help of our dashboard, one can create novel counterfactuals to assess the robustness of NLI models or use them to augment training sets.

\section{Preliminaries}
Before introducing our approach, we first go through some preliminaries. We briefly describe the NLI task and a tool called Data Maps.

\paragraph{Natural Language Inference (NLI).}
We employ our dashboard for the NLI task. The task is to determine whether a \defn{premise} \emph{entails}, \emph{contradicts}, or is \emph{neutral} to a \defn{hypothesis} \citep{condoravdi-etal-2003-entailment, 10.1007/11736790_9, bowman-etal-2015-large}. As with many other NLP tasks, neural NLI models have been shown to rely on spurious correlations \citep{gardner-etal-2021-competency}. For example, they often predict \emph{contradiction} when the hypothesis contains the word ``\emph{not}''. To obtain some hints on whether a model is relying on spurious correlations, we use data maps, which we describe next.

\paragraph{Data Maps.}
\citet{swayamdipta-etal-2020-dataset} propose a tool called \defn{Data Maps} to diagnose the characteristics of datasets with respect to a model's behavior during training. They propose two axes to locate each training data point in two dimensions. First, the \defn{confidence} is defined as the average probability that the model assigns to the \emph{true} label throughout training checkpoints, and second, the \defn{variability} of its confidence across the checkpoints. They identify three different regions in data maps: i) a region consisting of data points where the model has high confidence with low variability, i.e., \defn{easy to learn} data points, ii) a region consisting of data points where the model's confidence on the true label fluctuates a lot (high variability), i.e., \defn{ambiguous} data points, and iii) a region where the model has low confidence on the true label with low variability, i.e., \defn{hard to learn} data points.

In this paper, we employ data maps at two stages. First, in \cref{sec:locate} we discuss how to use data maps to locate DCCs. Second, we incorporate data maps in our interactive dashboard and further provide estimates of the location of newly created data points in the data map. Such an estimate gives early feedback to the user on how challenging it could be for the model to predict the label of the generated counterfactual. The user can then act on this feedback by revising the counterfactual.

\section{Data-Constrained Counterfactuals} \label{sec:locate}
In this work, we propose to start with finding existing counterfactuals in datasets. We will later use these counterfactuals in our dashboard \cref{sec:dashboard} to find spurious features and generate new data points. 

A data-constrained counterfactual is a data point that shares some features with other data points in the dataset but has a different label. Further, we want to make sure that the model is sensitive to the spurious correlation. Therefore, it \emph{should not} be easy for the model to label a DCC correctly. We provide the following formal definition of data-constrained counterfactuals.   

\begin{defin} \label{def:dcc}
A data point is a data-constrained counterfactual (DCC) when it satisfies two conditions: i) there exists other data points in the training set that are \emph{similar} to this data point but have a different label, and ii) it is not easy for the model to label it correctly, i.e., it falls into either the hard-to-learn or the ambiguous region in the data map.  
\end{defin}
This definition relies on a notion of similarity; thus, to identify DCCs we need to provide a similarity metric between data points. Following \citet{liu-etal-2022-wanli}, we define the similarity between data points as the cosine similarity between the \cls embedding of data points given by the underlying model. This will give us a tractable measure to find similar data points in large datasets without any manual inspection of the data. 

A caveat to \Cref{def:dcc} is that many data points in the hard-to-learn region have been found to be mislabeled \citep{swayamdipta-etal-2020-dataset}. To filter out samples that are likely to be mislabeled, we only select samples that have multiple annotations, where a large\footnote{$\geq 75\%$, as most multiple-annotated SNLI samples have four label annotations.} majority of annotators agree on the label.

\section{Visual Interactive Dashboard} \label{sec:dashboard}

In this section, we describe the tasks that users can perform during the interactive counterfactual generation process. We categorize these tasks using the \textit{explAIner} framework~\cite{SpinnerEtAl2020}. % in the human-computer interaction literature.
 
\subsection{Understanding}
First, the user is provided with enough information and supporting visuals to \emph{understand} the DCC that is being selected. This involves two tasks, explained below.
% \paragraph{\ti{1} Understanding the selected DCC.}
\paragraph{\ui{1} Data-Constrained Counterfactuals.}
The premise, hypothesis, and label of the DCC are shown to the user. This ensures that the user can get an initial understanding of the example and the annotators' reasoning.

% \paragraph{\ti{2} Understanding its location in the datamap.}
\paragraph{\ui{2} Data Map Contextualization.} 
The ground truth labels of the selected DCCs are inherently hard for the model to predict (see \Cref{def:dcc}). Therefore, it is helpful for the user to understand how the model reasons about the data point, i.e., how likely it is that the model predicts the correct label (confidence) and how often its prediction varies across different checkpoints (variance). To this end, we locate the selected data point in the data map (see \cref{fig:dashboard} black data point) and visualize the data map in our dashboard.
\subsection{Diagnosing}
Next, we aim to diagnose the reason that the DCC ends up being a counterfactual. As mentioned earlier, we aim to find features that correlate spuriously with the label. To find common features between the DCC and other data points in the sentence, we visualize similarities and differences between the DCC and other data points in the dataset. This involves performing two tasks explained below.
% \paragraph{\ti{3} Diagnose the similarities between the nearest neighbors.}
\paragraph{\di{1} Nearest Neighbor Similarities. }
We show two different sets of sentences in separate boxes and locate both sets in the data map. First, the set of sentences that are most similar to the DCC (in the blue box in \Cref{fig:dashboard}). By definition (\cref{def:dcc}), the most similar data points will have a \emph{different} label compared to the selected data point. By comparing the DCC with the most similar data points, one might be able to find structures or patterns that are shared between the two. Those can be features that spuriously correlate with the label. Second, we depict the set of most similar data points with \emph{the same} label as the data point (orange box in \Cref{fig:dashboard}). There might be more than one DCC breaking the spurious correlation in the dataset, and visualizing similar data points with \emph{the same} label can help the user discover such examples and their similarities to the DCC. In sum, investigating the similarities and differences between these two sets will help the user to diagnose potential spurious features that are shared between the sets and correlate with the label. \looseness=-1

\paragraph{\di{2} NLI Label Differences.}
We are interested to determine which sentences in the training dataset may have influenced the DCC being mislabeled. For very similar samples, the labels of the nearest \cls neighbors are a strong indication of what the model would predict for the seed sample. Therefore, we visualize the label of nearest neighbors in the data map using three distinct shapes. 

\subsection{Refining}
We will assist the user to create counterfactuals similar to the DCC, by pulling suggestions from GPT-3. The user can then refine the suggestion based on the feedback from the model.  
\paragraph{\ri{1} GPT-3 Suggestions.}
Following \citep{liu-etal-2022-wanli}, we use similar sentences with \emph{the same} label to prompt GPT-3 and create suggestions. Ideally, GPT-3 would find the reasoning pattern and generate a valid counterfactual sentence. However, as one can imagine GPT-3 might fail to generate a valuable sample for several different reasons, e.g., it might generate an example that is semantically close to the DCC but the reasoning is not aligned with the DCC. Another reason would be to generate an example that is easy for the model to learn. Therefore, we ask the user to refine this new example before adding it to the dataset.
% \paragraph{\ti{6} Get an estimate of datamap location.}
\paragraph{\ri{2} Data Map Location Estimation. }
One of the common errors with GPT-3 suggestions is that the suggestion might be easy for the model to learn. To filter those suggestions, after labeling the example, the user can request an estimate of the data map location. To ensure low latency for estimating the data map location of new examples, we do \emph{not} retrain the model. Instead, we receive the label from the user and use the saved checkpoints to measure the confidence of the model on the true label and its variance across the checkpoints. The user can then iteratively refine the example if it ends up in the easy-to-learn region. 

\section{Experimental Setup} 
The components of the dashboard are shown in \cref{fig:dashboard}. The filtering of potential DCCs described in section \cref{sec:locate} was performed on the SNLI dataset \citep{bowman-etal-2015-large}.\footnote{DCC definition relies on having a large set of samples with multiple annotations, which is available in SNLI dataset.} We compute the nearest neighbors of the DCCs according to the cosine similarity between the \cls embeddings extracted from a \textsc{RoBERTa}-large model \citep{https://doi.org/10.48550/arxiv.1907.11692} trained on SNLI. The data map was generated following \citep{swayamdipta-etal-2020-dataset}, where six end-of-epoch checkpoints of the SNLI \textsc{RoBERTa}-large model were used to estimate the data map location.

\begin{figure}
    \centering
    \includegraphics[width=0.45\textwidth]{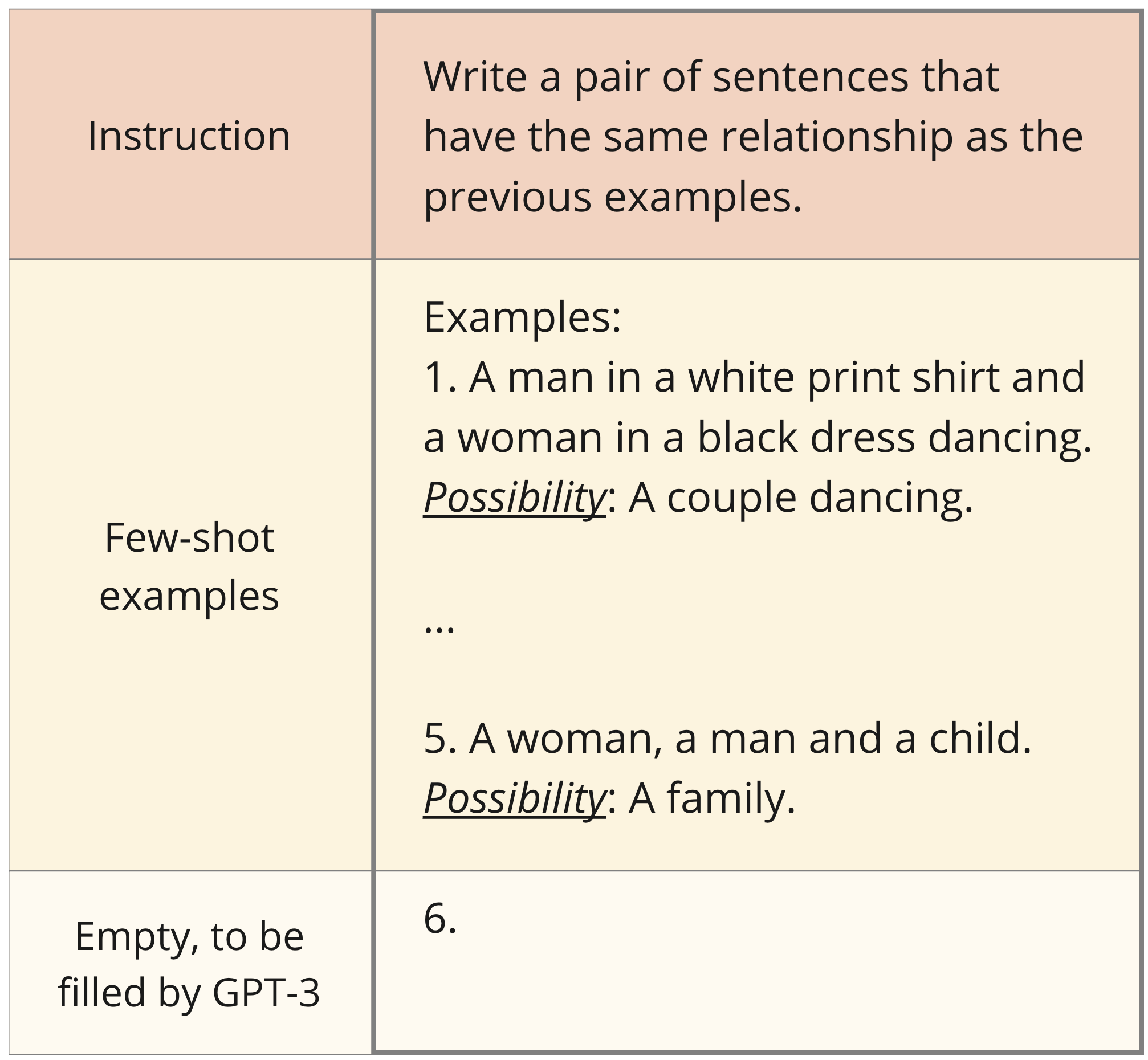}
    \caption{Example of GPT-3 few-shot prompting.  The few-shot examples are the nearest neighbors with the same label as the DCC, ordered in increasing DCC similarity, and finally, the DCC. The word setting the premise in context with the hypothesis can be either \textit{Implication, Possibility}, or \textit{Contradiction}, depending on whether the DCC is labeled \textit{entailment, neutral}, or \textit{contradiction}.}
    \label{fig:promting}
\end{figure}

Suggestions are generated by few-shot prompting the GPT-3 \texttt{davinci-003} model \citep{gpt3} using four nearest neighbors to the DCC with the same label, exactly following \citep{liu-etal-2022-wanli}, as they argue that the model may employ similar reasoning for such nearest neighbors. An example of such a prompt is shown and described in \cref{fig:promting}.%\afra{should "examples:" be in the instruction? I would move that down.}\robin{done}

\section{Findings}
By interacting with the dashboard ourselves in multiple sessions, we find interesting patterns and many DCC instances following those patterns. In this section, we provide a categorization of our findings. 

We find three high-level features that correlate spuriously with the label, which we name: Semantic Relevance, Logical Fallacies, and Bias.
%\afra{we should check to see if we are referring to these three categories + one artifact consistently}
%\robin{seems fine. Either we are referring to 4 categories or the 3 categories containing spurious correlations (i.e., not the artifacts one).}
Furthermore, we discovered another category that surfaces artifacts in the data collection procedure. Next, we will go through and explain each category, and further, provide some examples. 

\subsection{Semantic Relevance}
We find many instances in the dataset where the hypothesis is the rephrased version of the premise. Clearly, in those cases, the gold label is entailment, e.g., \ref{ex:rel-pos}.
\ex. \label{ex:rel-pos} {\textit{Premise:} A man in blue shorts and a t-shirt is slicing tomatoes on a dining table. \emph{\textbf{entails}} \\
\textit{Hypothesis:} A man prepares tomatoes by slicing them at the table.
}

However, if such examples dominate the dataset, a trained model might associate the entailment label to \emph{any} premise and hypothesis that are semantically relevant to each other. The semantic relevance is a spurious feature, as one can imagine counterfactual examples where the premise and hypothesis are semantically related but the premise \emph{does not} entail the hypothesis. One DCC that contains this feature is \ref{ex:rel-neg}.
\ex. \label{ex:rel-neg} {\textit{Premise:} A large group of people are walking towards something, and most of them have backpacks. \emph{\textbf{is neutral to}} \\
\textit{Hypothesis:} A group of people move toward something that \emph{requires} the use of a backpack.
}

In this example, while premise and hypothesis are semantically related, the word ``requires'' in the hypothesis makes the hypothesis to \textit{neutral} to the premise, while the NLI model predicts the \emph{entailment} label.

\subsection{Logical Fallacies}
Another common pattern we find in the dataset is hypotheses that become \emph{neutral} to the premise by mentioning extra details. 
\ex. \label{ex:common-pos} {\textit{Premise:} A woman in a black dress and flat shoes holds her head as she waits to cross the street. \emph{\textbf{is neutral to}} \\
\textit{Hypothesis:} The woman is carrying a purse.
}

For example, in \ref{ex:common-pos} the premise describes the appearance of a woman but does not mention anything about whether she is carrying a purse. Therefore, the hypothesis is referring to an extra piece of information that was not mentioned in the premise and thus, is neutral to it. If such examples dominate the dataset, a trained model might associate any extra information in the hypothesis with a neutral label. However, in some scenarios, the presence of logical clues in the premise will result in a different label. Such a DCC in the data is shown in \ref{ex:common-neg}.
\ex. \label{ex:common-neg} {\textit{Premise:} A man wearing \emph{only} red pants does a trick on a ladder. \emph{\textbf{contradicts}} \\
\textit{Hypothesis:} The man is wearing a black shirt.
}

In this example, while the premise does not directly talk about whether the man is wearing a black shirt or not, the word ``only'' indicates that the hypothesis is in fact, false. However, the NLI model predicts the \emph{neutral} label. 
\subsection{Biases}
As with many other datasets, NLI datasets contain instances of different sorts of biases. Gender stereotypes in professions are one example. 
\ex. \label{ex:bias-gender} {\textit{Premise:} A wrestler is jumping off of the ring to hit his competitor. \emph{\textbf{is neutral to}} \\
\textit{Hypothesis:} Two men are competing in a wrestling match.
}

In the above example \ref{ex:bias-gender}, while there is no mention of the gender of wrestlers in the premise, the model predicts that the hypothesis \emph{entails} the premise. This could be due to the fact that wrestling is stereotypically associated with men. 
\ex. \label{ex:bias-family} {\textit{Premise:} A woman, man, and two children. \emph{\textbf{is neutral to}} \\
\textit{Hypothesis:} A family.
}

Another example is \ref{ex:bias-family}, where we do not know the woman, man, and two children that the premise is describing are in fact a family. However, the model predicts \emph{entailment} as the label for this example.

\subsection{Artifacts}
The last category of examples is the existing artifacts in the dataset that surfaces in our dashboard. We find several examples where the hypothesis is completely irrelevant to the hypothesis, but their labels are inconsistent and often wrong.

\ex. \label{ex:art-con} {\textit{Premise:} A child and woman exchange glances. \emph{\textbf{contradicts}} \\
\textit{Hypothesis:} a bird was on rocks.
}

\ex. \label{ex:art-ent} {\textit{Premise:} A little child playing in water with a hose. \emph{\textbf{entails}} \\
\textit{Hypothesis:} a bird was on rocks.
}

While both examples \ref{ex:art-con} and \ref{ex:art-ent} should have a \emph{neutral} label, they are labeled as contradiction and entailment.

\section{Discussion}

The visual interactive dashboard for diagnosing spurious correlations and counterfactual generation can open up research opportunities in the following domains:

\paragraph{Bi-Directional Explanation of Reasoning Patterns.}
Our dashboard opens up a possibility for efficient collaboration between humans and AI. AI can help humans to find and group similar structures. As can be seen in our dashboard, similarities in the representation space of NLI models often capture similar structures. On the other hand, humans can explain the reasoning to AI. This can happen by generating new examples that follow a particular line of reasoning that is challenging for the AI model to learn, which can result in improving AI models. 
 
\paragraph{Diversifying Training Data based on DCC.}
 Receiving an estimate of model confidence during refinement (\ri{2}) enables the user to understand and pinpoint the patterns that pose a challenge to the model. Given the user has established such an understanding, they can produce samples that target a specific reasoning pattern. Further, GPT-3 suggestions assist the user by providing a diverse set of examples that follow the desired reasoning pattern. 
Therefore, the process allows us to augment potentially biased training datasets with a large, diverse set of counterfactuals. Conducting a thorough investigation, including large-scale expert annotation, model-retraining, and benchmarking is still required and will be part of future work.

\paragraph{Towards more Robust NLI Models.}
The counterfactual samples generated using our dashboard can be used as adversarial test suites for evaluating existing models. As a proof-of-concept, we generate a small set of such samples through our dashboard to evaluate a model trained on WaNLI data \citep{liu-etal-2022-wanli},\footnote{We used the \texttt{roberta-large-wanli} model released on huggingface \citep{wolf-etal-2020-transformers}.} which itself was trained to be more robust and results in state-of-the-art results on various NLI test suites. The WaNLI model only achieves an accuracy of around 30\% on our generations. This hints at the potential of our proposed annotation workflow for generating test suites to evaluate the robustness of NLI models.

\section{Related Work}
Other tools have been proposed for counterfactual generation. For example, \textsc{Polyjuice} \citep{wu-etal-2021-polyjuice} introduces an automated counterfactual generation based on minimal edits. Counterfactuals are created using a fixed set of control codes to edit the existing sentences in the dataset. 

Further, systems have been developed for mixed-initiative adversarial sample generation. ANLI \citep{nie-etal-2020-adversarial} introduces an adversarial sample generation framework, where annotators are tasked to write hypotheses that may fool the model for a given context (i.e., a premise and label). Following a similar framework, Dynabench \cite{kiela-etal-2021-dynabench} presents a more general-purpose dashboard for adversarial generation using model predictions and explanations as feedback to the user.\looseness=-1

Compared to the methods described above, our proposed approach aims to first diagnose potential spurious correlations through DCCs, and then generate counterfactuals based on the found spurious correlations via prompting large language models. Furthermore, our dashboard gives feedback to users during the refinement stage by providing them with data map estimates for newly generated counterfactuals.\looseness=-1

% Further, our system makes use of a data map location estimation as arguably richer diagnostic model feedback.

\section{Conclusion}
We present a dashboard to diagnose spurious correlations and artifacts that an NLI model may have acquired during training. We first provide a systematic approach to find data-constrained counterfactuals, i.e., existing counterfactuals in the dataset. We then feed the DCCs to our dashboard, where we contextualize them in the data map and also highlight the most similar data points in the dataset. By investigating similarities and differences between the data points, we were able to diagnose several spurious correlations, which we categorize into three different groups and a category of artifacts. Furthermore, we incorporate GPT-3 suggestions to allow for effective and diverse model-in-the-loop adversarial data generation. Therefore, our dashboard opens up future work on adversarial test suite generation and counterfactual augmentation. 

\section*{Acknowledgements}
We would like to thank the anonymous reviewers for their constructive and thorough feedback as well as Frederic Boesel and Steven H. Wang for their contributions in the early stages of the project. We would also like to thank Anej Svete for his helpful comments on the final version of the paper. This work was funded by the ETH AI Center.

\section*{Ethics Statement}
The authors foresee no ethical concerns with the research presented in this paper.

\bibliographystyle{acl_natbib}
\bibliography{custom}

\end{document}